\documentclass[a4paper,twoside]{article}

\usepackage{epsfig}
\usepackage{subcaption}
\usepackage{calc}
\usepackage{amssymb}
\usepackage{amstext}
\usepackage{amsmath}
\usepackage{amsthm}
\usepackage{multicol}
\usepackage{pslatex}
\usepackage{apalike}
\usepackage{booktabs}
\usepackage{multirow}
\usepackage{SCITEPRESS}     

\usepackage{todonotes}

\begin{document}

\title{Towards Multi-Agent Reinforcement Learning using \\ Quantum Boltzmann Machines}

\author{\authorname{
Tobias Müller,
Christoph Roch,
Kyrill Schmid and
Philipp Altmann}
\affiliation{Mobile and Distributed Systems Group, LMU Munich, Germany}
\email{tobias.mueller1@campus.lmu.de, \{christoph.roch, kyrill.schmid, philipp.altmann\}@ifi.lmu.de}
}


\abstract{Reinforcement learning has driven impressive advances in machine learning. Simultaneously, quantum-enhanced machine learning algorithms using quantum annealing underlie heavy developments. Recently, a multi-agent reinforcement learning (MARL) architecture combining both paradigms has been proposed. This novel algorithm, which utilizes Quantum Boltzmann Machines (QBMs) for Q-value approximation has outperformed regular deep reinforcement learning in terms of time-steps needed to converge. However, this algorithm was restricted to single-agent and small 2x2 multi-agent grid domains. In this work, we propose an extension to the original concept in order to solve more challenging problems. Similar to classic DQNs, we add an experience replay buffer and use different networks for approximating the target and policy values. The experimental results show that learning becomes more stable and enables agents to find optimal policies in grid-domains with higher complexity. Additionally, we assess how parameter sharing influences the agents' behavior in multi-agent domains. Quantum sampling proves to be a promising method for reinforcement learning tasks, but is currently limited by the Quantum Processing Unit (QPU) size and therefore by the size of the input and Boltzmann machine.}

\onecolumn \maketitle \normalsize \setcounter{footnote}{0} \vfill

\section{Introduction}
Recently, adiabatic quantum computing has proven to be a useful extension to machine learning tasks \cite{benedetti18,biamonte17,li18,neukart17}. Especially hard computational tasks with high data volume and dimensionality have benefitted from the possibility of using quantum devices with manufactured spins to speed-up computational bottlenecks \cite{neven08,rebentrost14,wiebe12}.

One specific type of machine learning is Reinforcement Learning (RL), where an interacting entity, called agent, aims to learn an optimal state-action policy through trial and error \cite{sutton18}. Reinforcement Learning has gained the public attention by defeating the 9-dan Go grandmaster \textit{Lee Sedol} \cite{silver16}, which has been thought to be impossible for a machine. In the latest years, reinforcement learning has seen many improvements, gained a large variety of application fields like economics \cite{charpentier20}, autonomous driving \cite{kiran20}, biology \cite{mahmud18} and even achieved superhuman performance in chip design \cite{mirhoseini20}. Reinforcement Learning has only seen quantum speed-ups for specials models \cite{levit17,neukart17,neukart17_2,paparo14}. Especially multi-agent domains have rarely been researched \cite{neumann20}.

Real-world reinforcement learning frameworks predominantly use deep neural networks (DNNs) as function approximators. Since DNNs are powerful - see the latest prominent example \textit{AlphaFold2} \cite{jumper20} - and can be run efficiently for large datasets on classical computers, deep reinforcement learning is able to tackle complex problems in large data spaces. Hence, there was little need for improvements.

However, since recent work has proved speed-ups for classical RL by leveraging quantum computing \cite{levit17,neumann20} and the application field gets more and more complex, it could be beneficial to explore quantum RL algorithms. These inspiring studies considered Boltzmann machines \cite{ackley85} as function approximator - instead of traditionally used DNNs. Boltzmann machines are stochastic neural networks, which are mainly avoided due to the fact, that their training times are exponential to the input size. Since finding the energy minimum of Boltzmann machines can be formulated as a "Quadratic Unconstrained Binary Optimization" (QUBO) problem, simulated annealing respectively quantum annealing is well suited to accelerate training time.

Nevertheless, the combination of RL and Boltzmann machines using (simulated) quantum annealing only worked properly for small single-agent environments and reached its limit at a simple $3 \times 3$ multi-agent domain. This work proposes an architecture inspired by DQNs \cite{mnih15} to enable more complex domains and stabilize learning by using \textit{experience replay buffer} and separating policy and target networks. We thoroughly evaluate the effects of these augmentations on learning.

Lately, an inspiring novel method to speed-up quantum reinforcement learning for large state and action spaces by proposing a combination of regular NNs and DBMs/QBMs, namely \textit{Deep Energy Based Networks} (DEBNs) was proposed \cite{jerbi20}. More specifically, these architectures are constructed with an input layer consisting of action and state units, which are connected with the first hidden layer through directed weights. This is followed by a single undirected stochastic layer. The remaining layers are linked with directed deterministic connections. Lastly, a final output layer returns the negative free energy $-F(s,a)$.

In contrast to QBMs, DEBNs therefore only comprise one stochastic layer, return an output similar to traditional deep neural networks and can be trained through backpropagation. DEBNs also use an experience replay buffer and separate the policy and target network. Additionally, they allow to trade off learning performance for efficiency of computation. Jerbi et al. briefly stated, that QBMs are applicable. Unfortunately, no numerical results were given for purely stochastic, energy-based QBM agents or domains with multiple agents. We aim to build on this.

Summarized, our contribution is three-fold:

\begin{itemize}
    \item We provide a Quantum Reinforcement Learning (Q-RL) framework, which stabilizes learning leading to more optimal policies
    \item Based on single- and multi-agent domains, we provide a thorough evaluation on the effects of an Experience Replay Buffer and an additional Target Network compared to traditional QBM agents
    \item Additionally, we demonstrate and discuss limitations to the concept
\end{itemize}

We first describe the preliminaries about reinforcement learning and quantum Boltzmann machines underlying the proposed architectures. Afterwards, the state-of-the-art algorithm and extensions made to it will be explained. We test and evaluate the approach and finally discuss restrictions and potential grounds for future work.
\section{Preliminaries}
This chapter describes the basics needed to understand our proposed architecture. First, reinforcement learning and the underlying Markov Decision Process will be explained followed by Boltzmann Machines and the process of quantum annealing.

\subsection{Reinforcement Learning}
We first describe Markov Decision Processes as the underlying problem formulation which is followed by an introduction to reinforcement learning in general. The subsequent sections specify independent and cooperative multi-agent reinforcement learning.

\paragraph{Markov Decision Processes} The problem formulation is based on the notion of Markov Decision Processes (MDP) \cite{puterman94}. MDPs are a class of sequential decision processes and described via the tuple $M = \langle S, A, P, R \rangle$, where

  \begin{itemize}
    \item $S$ is a finite set of states and $s_t \in S$ the state of the MDP at time step $t$. 
    \item $A$ is the set of actions and $a_t \in A$ the action the MDP takes at time step $t$.
    \item $P(s_{t+1}|s_t,a_t)$ is the probability transition function. It describes the transition that occurs when action $a_t$ is executed in state $s_t$. The resulting state $s_{t+1}$ is chosen according to $P$.
    \item $R(s_t,a_t)$ is the reward, when the MDP takes action $a_t$ in state $s_t$. We assume $R(s_t,a_t) \in \mathbb{R}$
   \end{itemize}
  
Consequently, the cost and transition function only depend on the current state and action of the system. Eventually, the MDP should find a policy $\pi : S \rightarrow A$ in the space of all possible policies $\Pi$, which maximizes the return $G_t$ at state $s_t$ over an infinite horizon via:
\begin{equation}
    G_t = \sum_{k=0}^\infty \gamma^k \cdot R(s_{t+k}, a_{t+k}), 
\end{equation}
with $\gamma \in [0,1]$ as the discount factor. This policy is called the optimal policy $\pi$.

\paragraph{Reinforcement Learning}\label{rl}
Model-free reinforcement learning \cite{strehl09} is considered to search the policy space $\Pi$ in order to find the optimal policy $\pi^*$. The interacting reinforcement learning agent executes an action $a_t$ for every time step $t \in [1,..]$ in the MDP environment. In model-free algorithms, the agent acts without any knowledge of the environment and the algorithm only keeps information of the value-function. Therefore, the agent knows its current state $s_t$ and the action space $A$, but neither the reward nor the next state $s_{t+1}$ of any action $a_t$ in any state $s_t$.\\
Consequently, the agent needs to learn from delayed rewards without having a model of the environment. A popular value-based approach to solve this problem is Q-learning \cite{peng04}. In this approach, the action-value function $Q^\pi: S x A\rightarrow \mathbb{R}, \pi \in \Pi$ describes the accumulated reward $Q^\pi(s_t, a_t)$ for an action $a_t$ in state $s_t$. The optimal Q-learning function $Q^*$ is approximated by starting from an initial guess for $Q$ and updating the function via:

\begin{equation}
\resizebox{1.0\hsize}{!}{$
    Q(s_t, a_t) \leftarrow  Q(s_t, a_t)+ \alpha [r_t+ \gamma \max_a Q(s_{t+1},a) - Q(s_t, a_t)]$}
\end{equation}

The learned Q-function will eventually converge to $Q^*$, which then implies an optimal policy. In the traditional experiments a deep neural network is used as a parameterized function approximator to calculate the optimal action for a given state.
\paragraph{Independent Multi-Agent Learning}\label{ma}
When multiple agents interact with the environment,
A fully cooperative multi-agent task can be described as a stochastic game G, defined as in \cite{foerster17} via the tuple $G = \langle S, A, P, R, Z, O, n, \gamma \rangle$, where:
  \begin{itemize}
    \item $S$ is a finite set of states. At each time step $t$, the environment has a true state $s_t \in S$.
    \item $A$ is the set of actions. At each time step $t$ each agent $ag$ simultaneously chooses an action $a^{ag} \in A$, forming a joint action $a \in A \equiv A^n$.
    \item $P(s_{t+1}|s_t,a_t)$ is the probability transition function as previously defined.
    \item $R(s_t,a_t)$ is the reward as previously defined. All agents share the same reward function.
    \item $Z$ is a set of observations of a partially or fully observable environment.
    \item $O(s,ag)$ is the observation function. Each agent draws observations $z \in Z$ according to $O(s,ag)$.
    \item $n$ is the number of agents identified by $ag \in AG \equiv \{1,...,n\}$.
    \item $\gamma \in [0,1)$ is the discount factor.
    \end{itemize}
In \textit{independent} multi-agent learning algorithms, each agent learns from its own action-observation history and is trained independently. This means, every agent simultaneously learns its own Q-function \cite{tan93}.

\begin{figure}
    \begin{centering}
    \includegraphics[height=0.8\linewidth, width=0.9\linewidth]{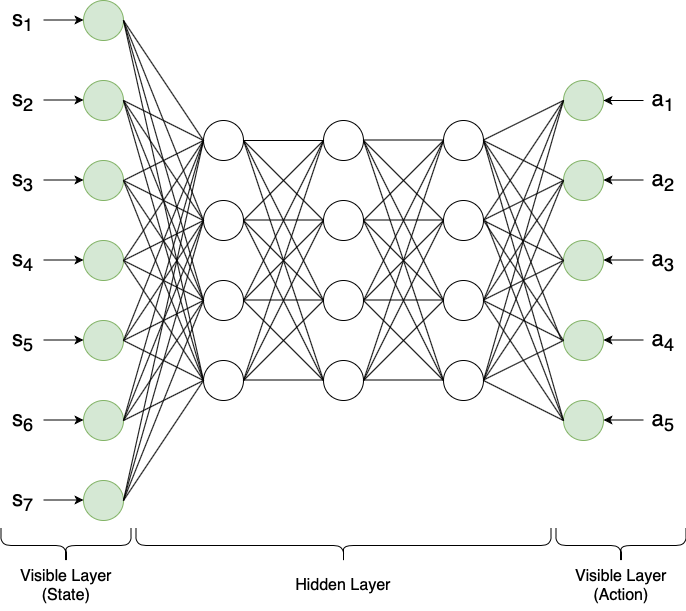}
    \caption{A Deep Quantum Boltzmann Machine with seven input state neurons and five input action neurons. The QBM additionally consists of three hidden layers with four neurons each. The state and action are given as fixed input and the configuration of the hidden neurons are sampled via (simulated) quantum annealing. The weights between two neurons are updated in the Q-learning step as described in section \ref{approach}.}
    \label{fig:dbm}
    \end{centering}
\end{figure}

\subsection{Boltzmann Machines}
The structure of a Boltzmann machine (BM) \cite{ackley85} is similar to Hopfield networks and can be described as a stochastic energy-based neural network. A traditional BM consists of a set of visible nodes $V$ and a set of hidden nodes $H$, where every node represents a binary random variable. The binary nodes are connected through real-valued, bidirected, weighted edges of the underlying undirected graph. The global energy configuration is generally given by the energy level of Hopfield networks. Since clamped BMs fix the assignment of the visible binary variables, these nodes are removed from the underlying graph and contribute as constant coefficients to the associated energy. Therefore the formula, which we aim to minimize, is given as the energy level of Hopfield networks with constant visible nodes:

\begin{equation}
    E(h) = - \sum_{i} w_{ii}v_{i} - \sum_{j} w_{jj}h_{j} - \sum_{i}\sum_{j}v_{i}w_{ij}h_{j},
\end{equation}
with $v_i$ as the visible nodes, $h_j$ as the hidden nodes and weights $w$. 

For this work, we implemented a Deep Boltzmann Machine (DBM) as trainable state-action approximator, which is constructed with multiple hidden layers, one visible input layer for the state and one visible action input layer. Finally, we modified the DBM to get a Quantum Boltzmann Machine (QBM), where qubits are associated to each node of the network instead of random binary variables \cite{crawford19,neumann20,levit17}. A visualization of a QBM for seven state neurons and five input action neurons can be seen in figure \ref{fig:dbm}. 
For any QBM with $v \in V$ and $h \in H$, the energy function is described by the quantum Hamiltonian $\mathcal{H}_{v}$:

\begin{equation}
\begin{split}
    \mathcal{H}_{v} = - \sum_{v,h} w_{vh}v\sigma^{z}_{h} - \sum_{v, v'} w_{vv'}vv' \\ - \sum_{h,h'}w_{hh'}\sigma^{z}_{h}\sigma^{z}_{h'} - \Gamma\sum_{h}\sigma^{x}_{h}
\end{split}
\end{equation} 

Furthermore, $\Gamma$ is the annealing parameter, while $\sigma^{z}_{i}$ and $\sigma{x}_{i}$ are spin-values of node $i$ in the $z-$ and $x-$ direction.
Because measuring the state of one direction destroys the state of the other, we follow the architecture of Neumann et al. (2020) \cite{neumann20} and replace all $\sigma^{x}_{i}$ by $\sigma^{z}$ by using replica stacking based on the Suzuki-Trotter expansion of the Hamiltonian $\mathcal{H}_{v}$. The BM is replicated $r$ times in total and connections between corresponding nodes in adjacent replicas are added. By this, we obtain a new effective Hamiltonian $\mathcal{H}^{eff}_{v=(s,a)}$ in its clamped version given by:
\begin{equation}
\begin{split}
    \mathcal{H}^{eff}_{v=(s,a)} = -\sum_{\substack{h \in H\\ h-s \ adj}} \sum^{r}_{k=1} \frac{w_{sh}}{r}\sigma_{h, k}
    -\sum_{\substack{h \in H\\ h-a \ adj}} \sum^{r}_{k=1} \frac{w_{ah}}{r}\sigma_{h,k} \\ 
     - \sum_{(h,h') \subseteq H}\sum^{r}_{k=1}\frac{w_{hh'}}{r}\sigma_{h,k}\sigma_{h',k} - \Gamma\sum_{h \in H}\sum^{r}_{k=0}\sigma_{h,k}\sigma_{h,k+1}
\end{split}
\end{equation}

For each evaluation of the Hamiltonian, we get a spin configuration $\hat{h}$. After $n_{reads}$ reads for a fixed combination of $s$ and $a$, we get a multi-set $\hat{h}_{s,a} = \{\hat{h}_{1}, ..., \hat{h}_{n_{reads}}\}$. We average over this multi-set to gain a single spin configuration $C_{\hat{h}_{s,a}}$, which will be used for updating the network. If a node is $+1$ or $-1$ depends on the global energy configuration:

\begin{equation}
    p_{node \ i=1} = \frac{1}{1+exp(- \frac{\Delta E_{i}}{T})}, 
\end{equation}
with $T$ as the current temperature.

Since the structure of Boltzmann Machines are inherent to Ising models, we sample spin values from the Boltzmann distribution by using simulated quantum annealing, which simulates the effect of transverse-field Ising model by slowly reducing the temperature or strength of the transverse field at finite temperature to the desired target value \cite{levit17}. As proven in \cite{satoshi08}, spin system defined by simulated quantum annealing converges to quantum Hamiltonian. Therefore it is straightforward to use simulated quantum annealing (SQA) to find a spin configuration for $h \in H$ - given $s \in S$ - which minimizes the free energy.
\section{Quantum Reinforcement Learning}\label{approach}
Recently, quantum reinforcement learning algorithms (QRL) using boltzmann machines and quantum annealing of single agent \cite{crawford19} and multi-agent domains \cite{neumann20} for learning grid-traversal policies have been proposed. 
Although, these architectures were able to learn optimal policies in less time steps compared to classic deep reinforcement learners (DRL), they could only be applied to single-agent or small multi-agent domains. Unfortunately, already $3 \times 3$ domains with 2 agents could not be solved optimally \cite{neumann20}. QRL seems to be unstable for more complex domains. We intuitively assume that BMs underlie similar instability problems as traditional neural networks. Hence, by correlations present in the sequence of observations and how small updates to the Q-values change the policy, data distribution and therefore the correlations between free energy $F(s_{n},a_{n})$ and target energy $F(s_{n+1}, a_{n+1})$. Inspired by Deep Q-Networks \cite{mnih15}, we propose to enhance the state-of-the-art architecture as described in section \ref{sota} by adding an experience replay buffer (see section \ref{erb}) to randomize over transitions and by separating the network calculating the policy and the network approximating the target value (see section \ref{target}) in order to reduce correlations with the target.

\subsection{State of the Art}\label{sota}
Traditionally, single-agent reinforcement learning using quantum annealing and QBMs is an adaption of Sallans and Hintons (2004) \cite{sallans04} RBM RL algorithm and structured as follows:
\paragraph{Initialization.}
The weights of the QBM are initialized by setting the weights using Gaussian zero-mean values with a standard deviation of 1.00. The topology of the hidden layers is set beforehand.
\paragraph{Policy.}
At the beginning of each episode, every agent is set randomly onto the grid and receives its corresponding observation. At each time step $t$, every agent $i$ independently chooses an action $a^{i}_{t}$ according to its policy $\pi^{i}_{t}$. To enable exploration, we implemented an $\epsilon$-greedy policy, where the agent acts random with probability $\epsilon$, which decreases by $\epsilon_{decay} = 0.0008$ with each training step until $\epsilon_{min} = 0.01$ is reached. When the agent follows its learned policy, we sweep across all possible actions and choose the action which maximizes the Q-value for state $s^{i}_{t}$. The Q-function of state $s$ and action $a$ is defined as the corresponding negative free-energy $-F$:
\begin{equation}
    Q(s,a) \approx -F(s,a) = -F(s,a;w),
\end{equation}
with $w$ as the vector of weights of a QBM and $F(s,a)$ as:

\begin{equation}
    F(s,a) = \langle H^{eff}_{v=(s,a)} \rangle - \frac{1}{\beta} P(s_{t+1}|s_{t},a_{t}) log P(s_{t+1}|s_{t},a_{t})
\end{equation}
Summarized, the agent acts via:
\begin{equation}
    \pi = \left\{\begin{array}{ll} random, & if \: p \geq \epsilon \\
             argmax \ Q(s,a), & if \: p < \epsilon \end{array}\right.
\end{equation}
for $a \in A$ and random variable $p$.

\paragraph{Weight Update.}
The environment returns a reward $r^{i}_{t+1}$ and second state $s^{i}_{t+1} \leftarrow a^{i}_{t}(s^{i}_{t})$ for each agent $i$. Based on this transition, the QBM is trained. The used update rules are an adaption of the state-action-reward-state-action (SARSA) rule by Rummery et al. (1994) \cite{rummery94} with negative free energy instead of Q-values \cite{levit17} defined as:
\begin{equation}
\begin{split}
    \Delta w^{vh} = \mu(r_{n}(s_{n},a_{n}) - \gamma F(s_{n+1},a_{n+1}) \\ + F(s_{n},a_{n}))v \langle \sigma^{z}_{h} \rangle
\end{split}
\end{equation}
\begin{equation}
\begin{split}
    \Delta w^{hh'} = \mu(r_{n}(s_{n},a_{n}) - \gamma F(s_{n+1},a_{n+1}) \\ + F(s_{n},a_{n}))v \langle \sigma^{z}_{h}\sigma^{z}_{h'}  \rangle,
\end{split}
\end{equation}
with $\gamma$ as the discount factor and $\mu$ as the learning rate. The free energy and configurations of the hidden neurons are gained by applying simulated quantum annealing respectively quantum annealing to the formulation of the effective Hamiltonian $H^{eff}_{v=(s,a)}$ as described in the previous section. At each episode, this process is repeated for a defined number of steps or until the episode ends.

\subsection{Experience Replay Buffer}\label{erb}
The first extension is a biologically inspired mechanism named experience replay \cite{mcclelland95,oneill10}. O'Neill et al. (2010) found, that the human brain stabilizes memory traces from short- to long-term memory by replaying memories during sleep and rest. The reactivation of brain-wide memory traces could underlie memory consolidation. Similar to the human brain, experience replay buffers used in deep Q-networks (DQN) store experienced transitions and provides randomized data during updating neural connections. Hence, correlations of observation sequences are removed and changes in the data distribution are smoothed. Furthermore, due to the random choice of training samples, one transition can be used multiple times to consolidate experiences.

To enable experience replay, at each time step $t$ we store the each agents' experience $e_{t} = (s_{t}, a_{t}, r_{t}, s_{t+1})$ in a data set $D_{t} = (e_{1},...,e_{t})$. For every training step, we randomly sample mini-batches from $D_{t}$ from which to Q-learning updates are performed.

This means, instead of updating the weights on state-action pairs as they occur, we store discovered data and perform training on random mini-batches from a pool of random transitions.

\subsection{Policy and Target Network}\label{target}
In order to perform a training step, it is necessary to calculate the policy value $F(s_{n}, a_{n})$ and target value $F(s_{n+1}, a_{n+1})$. Currently, policies and target values are approximated by the same network. Consequently, Q-values and target values are highly correlated. Small updates to Q-values may significantly change the policy, data distribution and target.

To counteract, we separate policy network calculating $F(s_{n},a_{n})$ from the target network approximating $F(s_{n+1},a_{n+1})$. Both networks are initialized similarly. The policy network is updated with every training step, whereas the target network is only periodically updated. Every $m$ steps, the weights of the policy network are simply adopted by the target network.

\subsection{Multi-Agent Quantum Reinforcement Learning}
In this work, we explore independent quantum learning in cooperative and non-cooperative settings. The explicit requirement for cooperation is communication \cite{binmore07}. We enable communication via parameter sharing as proposed by Foerster et al. (2016) \cite{foerster16}. In this case, every agents' transition is stored in a centralized experience replay buffer and only one BM is trained. Each agent receives its own observation and the centralised network approximates the agents' Q-value independently. Whereas in non-cooperative settings, every agent keeps and updates its own BM solely with its own experiences without any information exchange. The policy and weight updates are performed as described in the previous section.
\section{Evaluation}\label{eval}
\subsection{Domain}
\begin{figure}[hpbt]
 \centering
  \subfloat[3x3 grid]{
   \label{fig:STP:a}
   \includegraphics[width=0.366\linewidth]{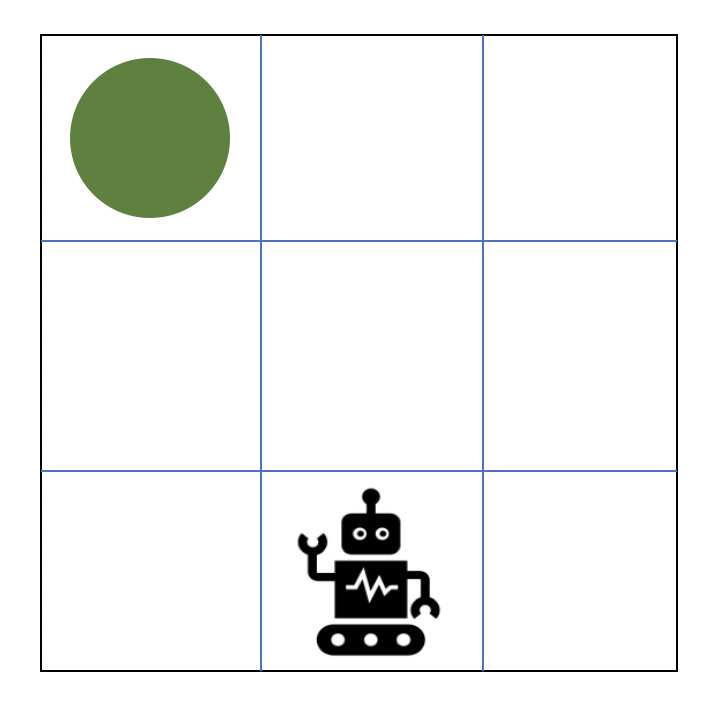}}
  \subfloat[5x3 grid]{
   \label{fig:STP:b}
   \includegraphics[width=0.6\linewidth]{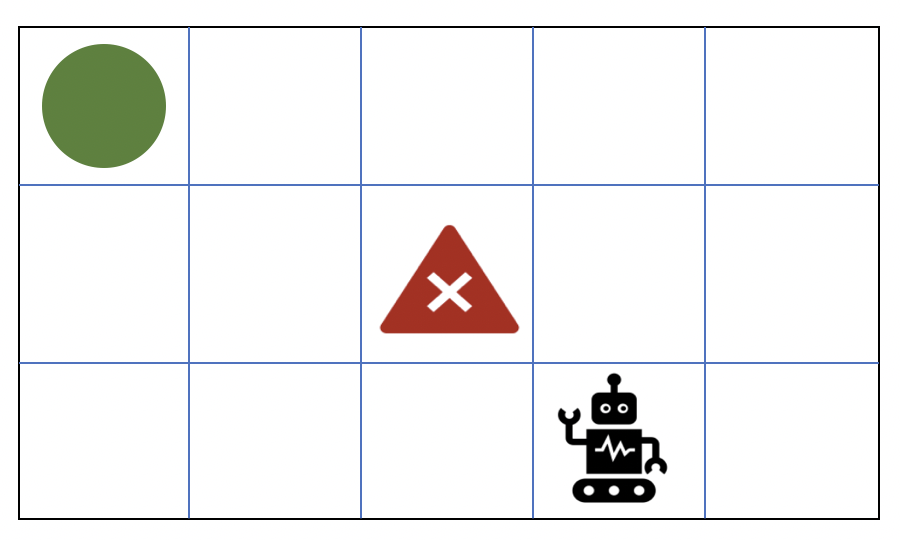}} 
 \caption[]{Example figures of two single-agent domains. Picture a) shows a $3 \times 3$ grid domain with one reward, whereas b) illustrates a bigger $5 \times 3$ grid domain with an additional penalty state.}
 \label{fig:single-agent-domain}
\end{figure}

To evaluate our approach, we implemented a discrete $n \times m$ multi-agent grid-world domain with $i$ deterministic rewards and $i$ agents.
At every time step $t$ each agent independently chooses an action from action space $A = \{up, \ down, \ left, \ right, \ stand \ still\}$ depending on the policy $\pi$. More specifically, the goal of every agent is to collect corresponding balls while avoiding obstacles (e.g. walls and borders) and penalty states (e.g. pits and others' balls). The environment size, number of agents, balls and obstacles can be easily modified. Reaching a target location is rewarded by a value of 220, whereas penalty states are penalized by -220 and an extra penalty of -10 is given for every needed step. An agent is done, when all its corresponding balls were collected. Consequently, we consider the domain as solved, when every agent is done. The main goal lies in efficiently navigating through the grid. Two example domains can be seen in figure \ref{fig:single-agent-domain}.

The starting position of all agents are chosen randomly at the beginning of each episode whereas the locations of their goals are fixed. The observation is one-hot-encoded and divided into two layers. One layer describes the agents' position and its goal and the other layer details the position of all other agents and their goals. This observation is issued as input for the algorithm. Therefore, the input shape is $n \times m \times 2$.
To asses the learned policies, we use the accumulated episode rewards as quality measure.

\subsection{Single-Agent Results}
\begin{figure*}[hpbt]
 \centering
  \subfloat[Training: Reward per Episode (3x3)]{
   \label{fig:STP:a}
   \includegraphics[width=0.45\linewidth]{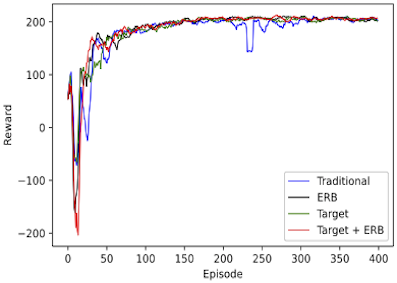}}
  \subfloat[Evaluation: Boxplot of Rewards (3x3)]{
   \label{fig:STP:b}
   \includegraphics[width=0.45\linewidth]{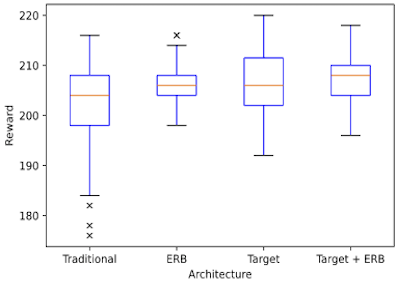}} \\
   \vspace{0.2cm}
  \subfloat[Training: Reward per Episode (5x3)]{
   \label{fig:STP:c}
   \includegraphics[width=0.45\linewidth]{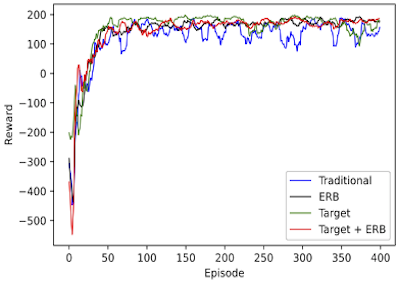}}
     \subfloat[Evaluation: Boxplot of Rewards (5x3)]{
   \label{fig:STp:c}
   \includegraphics[width=0.45\linewidth]{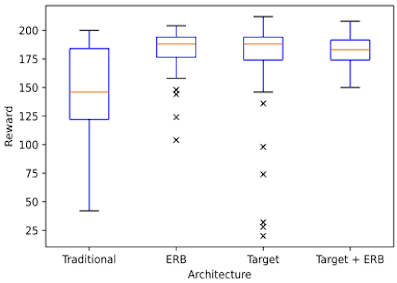}}
 \caption[]{Performance of a single agent with different architectures. a) Shows the gained reward per episode on a $3 \times 3$ domain of different architectures, whereas b) displays the corresponding achieved rewards of the learned policy on 400 test episodes. c) illustrates the reward of the same experiment on a $5 \times 3$ domain and d) the corresponding learned policy of test episodes.}
 \label{fig:eval3x3}
\end{figure*}

First, we evaluate how adding an experience replay buffer (ERB) and separating policy and target network influences the learning process and performance of a single agent. We started by running the traditional Q-RL algorithm as proposed by Neumann et al. (2020) \cite{neumann20} including their parameter setting. Then, we only added an experience replay buffer (ERB) respectively solely the target network. Finally, we extended the original algorithm with a combination of both, an ERB and target network. The resulting rewards on running all four architecture on the $3 \times 3$ domain (see figure  \ref{fig:single-agent-domain}) can be seen in figure \ref{fig:eval3x3} a) and the corresponding learned policy in figure \ref{fig:eval3x3} b). All graphs have been averaged  over ten runs. The traditional Q-RL agent without any extensions (blue line) learns unstable with occasional high swings down to -1700 and -1000 reward points. Extended versions seem to be show less outliers. This observation gets more evident, when conducting the same experiment on a bigger $5 \times 3$ environment. As seen in figure \ref{fig:eval3x3} c) - d) the achieved rewards of non-extended agents (blue) collapses frequently. The ERB (black) respectively target network (green) alone stabilize learning, but the combination of both (red) yields smoothest training curve. Hence, these enhancements are getting more important with bigger state space and more complex environments.

After training, we evaluate the resulting policies for 100 episodes without further training. The average rewards of ten test-runs on the $3 \times 3$ domain can be seen in figure \ref{fig:eval3x3} b). As already described, an agent is rewarded +220 points for reaching its goal and -10 for each taken step. So, when considering an optimal policy, the agent would be awarded +190 for the $3 \times 3$ domain (respectively +170 for $5 \times 3$) if the agent is spawned furthest from its goal and +220 for the best starting position. Assuming the starting positions over all episodes are distributed evenly, the optimal median reward would be at +205 for the $3 \times 3$ domain and +195 for the $5 \times 3$ environment.

The traditional QBM agent shows multiple outliers and a higher spread of rewards throughout the evaluation episodes compared to the other architectures. As it can be seen, adding only one of the extensions leads to a better median reward and a seemingly optimal policy is gained through a combination of both. Again, this observation gets more distinct with bigger domains, see figure \ref{fig:eval3x3} d). Even though ERB or target network alone significantly enhance the median reward, the plots still show outliers. The combined architecture is free of outliers with less interquartile range and lower overall span indicating reduced variance of training performance and nearly optimal policy. In summary, alleviating data correlation and the problems non-stationary distributions by randomly sampling previous transitions and separating target and policy network increases stable learning leading to robust and more optimal policies. Comparing the results for $3 \times 3$ with the $3 \times 5$ gridworld, a correlation of impact through the extensions and input size can be suspected.

\subsection{Multi-Agent Results}
\begin{figure}
  \centering
  \begin{tabular}[b]{c}
     \label{fig:STP:a}
    \includegraphics[width=0.9\linewidth]{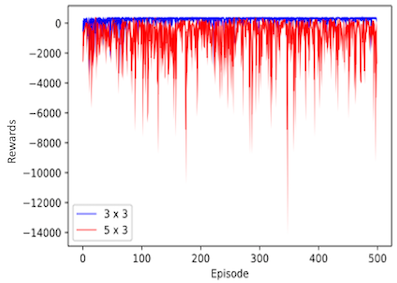} \\
    \small a) Learning Process on Both Domains
  \end{tabular} \qquad
  \begin{tabular}[b]{c}
     \label{fig:STP:b}
    \includegraphics[width=0.87\linewidth]{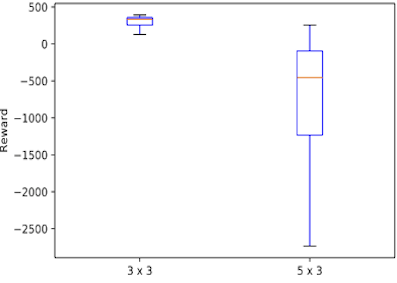} \\
    \small b) Learned Policy on Both Domains
  \end{tabular}
  \caption{Performance of two agents on the $3 \times 3$ (blue) respectively $3 \times 5$ (red) domain. Figure a) shows the learning process over 500 episodes, whereas figure b) displays the learned policy over 100 testing episodes.}
  \label{fig:eval_multi}
\end{figure}

Traditional Q-RL was limited so $2 \times 2$ multi-agent domains and bigger domains could not be solved rationally \cite{neumann20}. This section explores, if the proposed architecture enables multi-agent reinforcement learning. We modify the known environments by adding one agent and one corresponding goal. If an agents picks up the others goal, it is penalized with -220. The averaged results over 10 runs can be seen in figure \ref{fig:eval_multi}.

The graphs suggest, that $3 \times 3$ domain (blue) can be solved in contrast to the bigger environment (red). Looking at figure \ref{fig:eval_multi} b), the median reward of the learned policy on the smaller domain is around +350, which is near optimum. Unfortunately, the bigger domain could not be solved with a median reward of -450. Additionally, the $5 \times 3$ learning curve does not seem to converge. Therefore, we can conclude, that it is possible to solve bigger domains with the proposed architecture, but Q-RL with ERB and extra target network still fails in somewhat larger multi-agent domains.

Lastly, we explore if the cooperation method of parameter sharing enhances quantum multi-agent reinforcement learning. With parameter sharing no explicit communication is necessary since only one centralized entity is trained and shared between the agents. More specifically, the experience of every agent is stored in a centralized ERB. At each training step, one QBM is trained with a randomized sample from the ERB similar to the single-agent case. Both agents use this network to independently calculate their Q-values based their observation. By this, we additionally smooth the data distribution hoping to achieve a more general policy and not two specific policies adjusted to particular observations.

\begin{figure}
  \centering
    \begin{tabular}[b]{c}
    \includegraphics[width=0.9\linewidth]{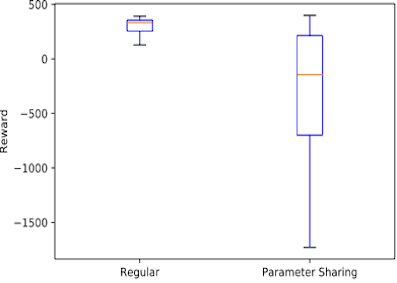} \\
    \small a) Learned Policy (3x3)
  \end{tabular}
    \begin{tabular}[b]{c}
    \includegraphics[width=0.9\linewidth]{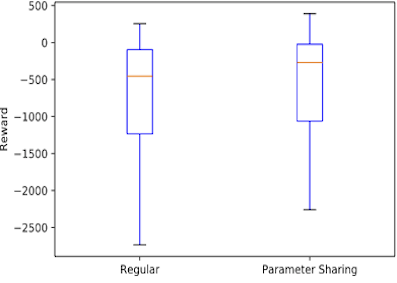} \\
    \small b) Learned Policy (5x3)
  \end{tabular}
  \caption{Performance of a two agents with and without parameter sharing. a) Shows the gained reward of the learned policy of 100 testing episodes on a $3 \times 3$ domain, whereas b) displays the same experiment on the bigger environment.}
  \label{fig:paramsharing}
\end{figure}

The results with and without parameter sharing are illustrated in figure \ref{fig:paramsharing}. Unfortunately, parameter sharing seems to have a negative effect on the small $3 \times 3$ domain. In this case, the agents seem to have learned a worse policy with this adaption. Rewards on the bigger environments have increased. However, the $5 \times 3$ domain can still not be considered solved. Hence, parameter sharing is sub-optimal for the evaluated use case.

The complexity of the task and size of the input did not increase, so this observation is counter-intuitive. Since the centralized entity is simultaneously learning two independent behaviors, it might be possible that in this case two independently optimal action-state probability distributions (as learned without parameter sharing) cancel out each other when learned together. To proof this assumption, more experiments must be conducted.

\section{Discussion}
In summary, adding an ERB and additional target network alleviates data correlation and the problem of non-stationary distribution resulting in stabilized learning and a more optimal policies. With the proposed architecture, we were able to solve bigger environments compared to traditional MARL using QBMs. However, this architecture is still limited to relatively small domains.

Even though it is possible to coordinate a single agent in the $5 \times 3$ domain and multiple agents in a smaller domain. The question remains why the $5 \times 3$ multi-agent domain fails. The QBM-agent receives an input of 15 neurons on $5 \times 3$ single-agent domain since only one input layer is needed. When adding more agents to the environment, there is another input layer necessary in order to distinguish between the acting agent and other opposing agents. Hence, the $3 \times 3$ multi-agent domain returns an observation size of 18 and bigger multi-agent domain of size 30. The input are considered in the QUBO formulation, which therefore increases. Hence, simulated quantum annealing is applied to a bigger formulation. A bigger formulation demands more qubits, which may limit the accuracy, variation and stability of the quantum annealing algorithm. This is only an assumption and needs to be examined more closely. Neumann et al. (2020) also already stated, that Q-RL is limited by the current Quantum Processing Unit (QPU) size. However, with the extension of an Experience Replay Buffer and Target Network, we are able to stabilize learning and therefore may reduce the needed QPU size compare to previous approaches. 

Quantum sampling has been proven to be a promising method to enhance reinforcement learning tasks to speed-up learning in relation to needed time steps \cite{neumann20}. Further work concerning the relation between QPU size and domain complexity (respectively state input) would needed to strictly determine current limitations.

\section*{\uppercase{Acknowledgements}}

This work was funded by the BMWi project PlanQK (01MK20005I).

\bibliographystyle{apalike}
{\small
\bibliography{sources/sources}
}

\end{document}